\documentclass[runningheads]{llncs}

\usepackage{eccv}

\usepackage{eccvabbrv}

\usepackage{graphicx}
\usepackage{multirow}
\usepackage{booktabs}
\usepackage{algorithm}
\usepackage{algpseudocode}
\usepackage{dsfont}
\usepackage{wrapfig}
\usepackage{amssymb}

\definecolor{softblue}{RGB}{40,80,150}  % muted academic blue

\usepackage[accsupp]{axessibility}  % Improves PDF readability for those with disabilities.

\usepackage[pagebackref,breaklinks,colorlinks,citecolor=eccvblue]{hyperref}
\usepackage{orcidlink}

\newcommand{\model}{ExeVRM\xspace}
\newcommand{\dataset}{ExeVR-53k\xspace}
\newcommand{\testset}{ExeVR-Bench\xspace}
\usepackage{xspace}

\usepackage[most]{tcolorbox}
\usepackage{xcolor}

\tcbuselibrary{breakable}

\begin{document}

% \title{Computer-Use Reward Modeling from Video}
\title{Video-Based Reward Modeling for \\Computer-Use Agents}

\titlerunning{Video-Based Reward Modeling for Computer-Use Agents}

\author{Linxin Song\inst{1} \and Jieyu Zhang\inst{2} \and Huanxin Sheng\inst{3} \and Taiwei Shi\inst{1} \and Gupta Rahul\inst{4} \and Yang Liu\inst{4} \and Ranjay Krishna\inst{2} \and Jian Kang\inst{3} \and Jieyu Zhao\inst{1}}

\authorrunning{L.~Song et al.}

\institute{$^{1}$University of Southern California \ $^{2}$University of Washington \\ $^{3}$MBZUAI \ $^{4}$Amazon AGI}

\maketitle

\begin{abstract}
Computer-use agents (CUAs) are becoming increasingly capable; however, it remains difficult to scale evaluation of whether a trajectory truly fulfills a user instruction. In this work, we study reward modeling from \emph{execution video}: a sequence of keyframes from an agent trajectory that is independent of the agent's internal reasoning or actions. Although video-execution modeling is method-agnostic, it presents key challenges, including highly redundant layouts and subtle, localized cues that determine success. We introduce Execution Video Reward 53k (\dataset), a dataset of 53k high-quality video--task--reward triplets. We further propose adversarial instruction translation to synthesize negative samples with step-level annotations. To enable learning from long, high-resolution execution videos, we design spatiotemporal token pruning, which removes homogeneous regions and persistent tokens while preserving decisive UI changes. Building on these components, we fine-tune an Execution Video Reward Model (\model) that takes only a user instruction and a video-execution sequence to predict task success. Our \model~8B achieves 84.7\% accuracy and 87.7\% recall on video-execution assessment, outperforming strong proprietary models such as GPT-5.2 and Gemini-3 Pro across Ubuntu, macOS, Windows, and Android, while providing more precise temporal attribution. These results show that video-execution reward modeling can serve as a scalable, model-agnostic evaluator for CUAs.
\end{abstract}

\noindent \texttt{Code:} \url{https://github.com/limenlp/ExeVRM}\\
\texttt{Model:} \url{https://huggingface.co/lime-nlp/ExeVRM-8B}\\
\texttt{Dataset:} \url{https://huggingface.co/datasets/lime-nlp/ExeVR-53K}

\section{Introduction}
Computer-use agents (CUA) are becoming a promising paradigm for general-purpose computer-using automation~\cite{uitars,uitars2,agents2,agi0,autoglm,jedi,agentic-lybic,Mano,mobile-agent-v3,gta-1,opencua,tianxi-action,ilie2025uipath,coact-1}. By operating directly on real-world interfaces such as desktops, browsers, and mobile apps, these agents can execute complex, multi-step tasks in the same environments humans use. Advances in multimodal foundation models have accelerated this direction~\cite{qwen25,qwen3,o3-o4,claude37,claude4,claude45,doubao15,gemini25,kimi-vl,nemotron-tool-n1,deitke2024molmopixmoopenweights,molmo2,wang2025videothinkersparkingthinkingvideos}, enabling agents to perceive visual layouts, interpret instructions, and interact across diverse platforms.

However, evaluating CUA remains challenging. Existing benchmarks often rely on handcrafted scripts or task-specific rules to verify completion~\cite{AER,simplifiedjudge,gao2025guishiftenhancingvlmbasedgui}, which limits scalability and transfer to new tasks and environments. A learned reward model is a more flexible alternative: given a user instruction and an execution trajectory, it judges whether the intended goal is achieved. To serve as a universal evaluator across heterogeneous CUA trajectories, the model should rely on a representation independent of internal reasoning/action formats (e.g., thoughts, tool calls, or code traces). We therefore focus on execution video, the observable sequence of interface states during interaction, which is method-agnostic and naturally comparable across different agent designs.

Building reward models from execution video introduces two key challenges. First, CUA trajectories are highly redundant: large interface regions (e.g., toolbars, backgrounds, and layout elements) remain nearly static across steps~\cite{showui,screenspot-pro,OSWorld}, while correctness often depends on subtle local changes such as cursor focus shifts, small text edits, or transient dialogs. The model must therefore suppress redundant content without missing decisive cues. Second, negative supervision is limited. Public computer-using datasets mainly emphasize successful or high-quality trajectories for agent training~\cite{opencua,liu2025scalecua,OSWorld}, and failure cases are rarely paired with explicit annotations of when and why trajectories diverge. This makes it difficult to build balanced, informative data for reward modeling.

To address these challenges, we construct a training dataset named Execution Video Reward 53k (\dataset), a large-scale corpus of execution video trajectories for reward modeling. We unify interaction data from multiple computer-using training datasets and agent rollouts by converting their logs into a consistent step-level video representation. Each trajectory is segmented into atomic interaction steps, and a representative key frame is extracted per step to capture the corresponding interface state. The resulting sequence forms a compact video summary that preserves the temporal progression of the task while remaining computationally manageable. To overcome the scarcity of negative supervision, we further introduce adversarial instruction translation, which synthesizes hard mismatched instruction--trajectory pairs within the same interface context. By generating plausible but semantically inconsistent instructions and identifying the step where the mismatch becomes evident, this process provides informative contrastive signals for training reward models.

To enable efficient learning from high--resolution, long--horizon execution videos, we propose a spatiotemporal token pruning strategy tailored to CUA execution video sequences. Spatial token pruning (STP) removes visually homogeneous regions that contribute little discriminative signal, such as large static backgrounds, while retaining localized UI elements that are more likely to encode task-relevant information. Temporal token pruning (TTP) further suppresses tokens that remain nearly unchanged across consecutive frames, allowing the model to focus on meaningful state transitions rather than repeated layout structure. Together, these mechanisms reduce redundancy while preserving subtle visual cues that determine correctness, making video-based reward modeling tractable and robust for CUA environments.

With \dataset and the proposed spatiotemporal pruning strategy, we fine-tune an Execution Video Reward Model (\model) based on Qwen3-VL~\cite{qwen3} to judge CUA task success directly from execution video demonstrations. The model takes as input a user instruction together with the corresponding execution video sequence and outputs a judgment of whether the trajectory satisfies the intended goal. By grounding its decision in the observed interface states over time, the model learns to associate subtle visual transitions with task success or failure. Experiments on \testset show that \model~8B reaches 84.7\% accuracy and 87.7\% recall, outperforming strong proprietary baselines (Seed-2.0 Pro~\cite{seed2}: 80.3/74.7; GPT-5.2~\cite{gpt52}: 75.0/66.5 in accuracy/recall) and open-weight models, while also achieving consistently better temporal attribution (higher tIoU) for localizing decisive error spans. Discussion and ablations further show that dense video context is critical compared with sparse screenshot-based evaluation, and that 720p inputs with STP+TTP improve completion judgment (notably recall) over 360p while maintaining tractable long-horizon training via favorable memory-latency trade-offs.

\section{Related Work}

\subsection{Reward Evaluation for Visual GUI Agents}
\label{subsec:gui_agents_reward}

Large multimodal models have rapidly advanced GUI agents that execute natural-language instructions. Recent work spans web~\cite{zhou2024webarena, deng2023mind2web,koh2024visualwebarena,scuba}, desktop~\cite{OSWorld, opencua,bonatti2025windows, macosworld, liu2025scalecua}, and mobile benchmarks~\cite{rawles2025androidworld}, alongside specialized GUI perception/action models~\cite{hong2024cogagent,showui, lu2024omniparser, uitars,uitars2,autoglm,jedi,Mano,mobile-agent-v3}. Yet reward evaluation remains a bottleneck. 

Most existing protocols still depend on hand-crafted rules and environment-specific parsers~\cite{OSWorld, bonatti2025windows, dai2025scubasalesforcecomputeruse,nnetnav, fang2025webevolverenhancingwebagent,pahuja2025explorerscalingexplorationdrivenweb}. Final-state checks are scalable but coarse~\cite{AER, simplifiedjudge, gao2025guishiftenhancingvlmbasedgui}; full-screenshot evaluation improves coverage~\cite{chen2025spabenchcomprehensivebenchmarksmartphone,lin2025cuarewardbenchbenchmarkevaluatingreward} but is memory-intensive and often forces downsampling, which can hide subtle yet decisive GUI cues. Methodologically, prior work has explored Outcome Reward Models (ORMs) that score only end results~\cite{qi2025webrltrainingllmweb, bai2024digirltraininginthewilddevicecontrol, yang2025zeroguiautomatingonlinegui}. Process Reward Models (PRMs) provide denser supervision~\cite{os_genesis, progrm, ui_genie_rm, chae2025webshepherd, sun2025seagentselfevolvingcomputeruse,wu2025osoraclecomprehensiveframeworkcrossplatform, wang2026gaiadataflywheeltraining, chen2025guishepherdreliableprocessreward, wang2026buildingautonomousguinavigation, webarbiter}, but usually require $O(n)$ step-wise inference and are vulnerable to error accumulation or reward hacking~\cite{zheng2026adaptivemilestonerewardgui}. Our \model instead performs holistic video-level judgment with first-failure localization, reducing step-wise tracing complexity; unlike GUI-critic-R1 and VAGEN, it relies only on external execution video, without agent-specific reasoning traces or tool calls~\cite{wanyan2025lookleapguicriticr1model,cui2026agenticrewardmodelingverifying}.

Negative supervision is another bottleneck because public computer-using data are dominated by successful trajectories~\cite{zhou2024webarena, deng2023mind2web, OSWorld}. Existing negative construction relies on passive failure collection~\cite{simplifiedjudge, chen2025spabenchcomprehensivebenchmarksmartphone}, expert annotation~\cite{lin2025cuarewardbenchbenchmarkevaluatingreward, chen2025guishepherdreliableprocessreward}, model annotation~\cite{qi2025webrltrainingllmweb, wanyan2025lookleapguicriticr1model}, or rule-based corruption~\cite{wu2025osoraclecomprehensiveframeworkcrossplatform,ui_genie_rm}. In contrast, we pair successful trajectories with semantically mismatched instructions to generate scalable hard negatives, while standardizing heterogeneous data sources into a unified representation~\cite{opencua, liu2025scalecua, OSWorld}.

\subsection{Efficient Video Understanding and Token Pruning}
\label{subsec:video_token_reduction}

Efficient token pruning is critical to enabling reward models to process GUI execution videos at scale. Prior methods for visual token pruning~\cite{bolya2023tokenmergingvitfaster, ryoo2025xgenmmvidblip3videoneed32,chen2024imageworth12tokens,shang2026llavaprumergeadaptivetokenreduction,prunevid,shao2025holitomholistictokenmerging, ye2025trethinkingtemporalsearch, chen2025letuslongeventtextunderstanding} are mainly developed for natural or egocentric images or videos and often target action recognition or step localization. In GUI trajectories, however, critical evidence is frequently subtle and transient across adjacent key frames, so current methods might remove semantically important UI cues. Meanwhile, large static regions within each frame introduce substantial token redundancy that inflates memory cost with limited reward signal~\cite{xu2026spatiotemporaltokenpruningefficient, huang2025guikvefficientguiagents}. However, methods tailored to GUI settings remain underexplored. Prior work study spatial-only dropout on single screenshot~\cite{showui, ouyang2026focusuiefficientuigrounding}, while we propose a more fine-grained spatiotemporal token pruning mechanism. Despite their spatiotemporal awareness, GUI-Pruner~\cite{xu2026spatiotemporaltokenpruningefficient} and GUI-KV~\cite{huang2025guikvefficientguiagents} are primarily designed for test-time memory saving during agent execution. In contrast, we adopt simpler spatial pruning and more robust temporal strategies to preserve tiny yet decision-critical visual evidence over long-horizon trajectories, which finally contributes to efficient reward modeling for outcome assessment and error attribution.

\section{Execution Video Reward Modeling}

\subsection{\dataset}

\begin{wrapfigure}{r}{0.52\textwidth}
  \centering
  \vspace{-2\baselineskip}
  \includegraphics[width=0.5\textwidth]{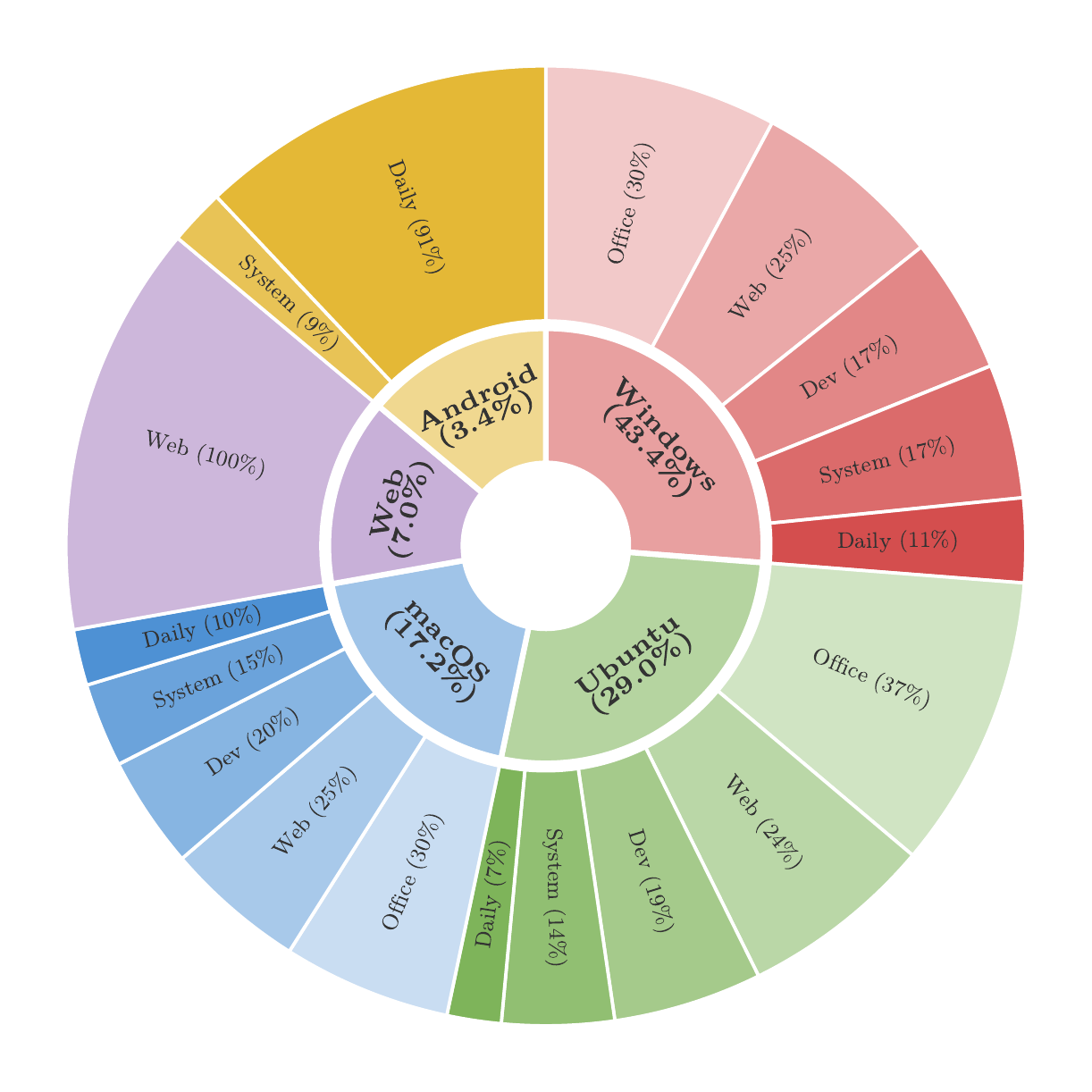}
  \caption{Task distribution of \dataset.}
  \label{fig:distribution}
  \vspace{-2\baselineskip}
\end{wrapfigure}

To address the data bottleneck in training reward models for computer use, we introduce Execution Video Reward 53k (\dataset), a training dataset of video trajectories curated from AgentNet~\cite{opencua}, ScaleCUA~\cite{liu2025scalecua}, and OSWorld~\cite{OSWorld}. Existing resources are either limited in scale or lack diversity in agent behaviors; \dataset bridges this gap by combining human-supervised demonstrations (AgentNet and ScaleCUA) with solutions generated by 30 diverse computer-using agents (OSWorld), spanning end-to-end and agentic paradigms. We summarize the resulting data composition and key statistics in \autoref{tab:dataset-stats}.

\begin{table}[t]
  \center
  \caption{Statistics of \dataset sources.}
  \label{tab:dataset-stats}
  \resizebox{\textwidth}{!}{%
    \setlength{\tabcolsep}{5pt}
    \begin{tabular}{lllll}
      \toprule
      \textbf{Dataset} & \textbf{Tasks} & \textbf{Trajectories} & \textbf{Trajectory Type} & \textbf{Accuracy} \\ \midrule
      AgentNet~\cite{opencua} & 23k & 23k & Human              & 50\% (w/ synthetic) \\
      ScaleCUA~\cite{liu2025scalecua} & 7k  & 7k  & Agent (human evaluation)              & 50\% (w/ synthetic) \\
      OSWorld~\cite{OSWorld}  & 361 & 23k & Agent (rule-based evaluation) & 32\%             \\ \bottomrule
    \end{tabular}%
  }
\end{table}

We build our training corpus by leveraging existing computer-using datasets: AgentNet, ScaleCUA, and OSWorld. We convert their interaction records into a unified step-level video representation. Concretely, we segment each recorded trajectory into step units and retrieve a representative key frame (a screenshot after an action) for each step to capture the corresponding UI state. We then concatenate these key frames in temporal order to form a compact video summary rendered at 1~FPS, coarse-grained progression of the interaction while keeping the input length manageable. \autoref{fig:distribution} summarizes the task composition of \dataset across sources, highlighting the diversity of operating environments and task categories covered in our collected trajectories.

\paragraph{OSWorld} \cite{OSWorld} focuses on realistic computer-using evaluation over open-domain web and desktop applications, including OS file I/O, multi-application workflows, and system-level operations. It contains 369 tasks (with 361 commonly evaluated and 8 optional Google Drive tasks). Unlike demonstration datasets, OSWorld does not provide fixed ground-truth trajectories; instead, each task is specified by environment initialization plus executable evaluation scripts, and computer-using agents generate trajectories during evaluation. In our pipeline, we treat rollouts from 30 different computer-using agents\cite{uitars,uitars2,agents2,agi0,autoglm,claude37,claude4,claude45,doubao15,gemini25,jedi,kimi-vl,agentic-lybic,Mano,mobile-agent-v3,o3-o4,gta-1,opencua,qwen25,tianxi-action,ilie2025uipath} on 361 tasks as trajectories and apply the same step segmentation and key-frame summarization procedure to obtain training-ready video sequences.

\paragraph{AgentNet} \cite{opencua} provides human demonstration trajectories spanning real-world desktop and web workflows. It covers four main domains with 11 finer-grained subdomains, including e-commerce, news \& entertainment, lifestyle, social media and communication, office tools, task management and collaboration, creative design and multimedia, development and engineering, knowledge discovery and research, data analysis, web tools and internet utilities, and operating systems and utilities. We use 22,625 human-labeled tasks from AgentNet. The trajectories are collected across multiple operating systems, including Windows (12K tasks), macOS (5K tasks), and Ubuntu (5K tasks), and are recorded as step-by-step human demonstrations.

\paragraph{ScaleCUA} \cite{liu2025scalecua} is a large-scale, GUI-centric dataset designed to support grounding and end-to-end computer-using learning across a diverse set of platforms. The paper defines three high-level task domains centered on GUI/action grounding, navigation and interaction, and trajectory-level task execution. The dataset additionally contains multiple interaction types, although exact per-domain statistics are not fully disclosed publicly in the dataset card. ScaleCUA spans Linux, macOS, Windows, Android, and the Web. Its data is produced via a hybrid pipeline: grounding examples are collected largely automatically and annotated with LLM assistance followed by human verification, while trajectory data relies on human interaction logs augmented with LLM-generated annotations.

\subsection{Adversarial Instruction Translation}
\begin{figure}[h]
  \centering
  \vspace{-5pt}
  \includegraphics[width=\textwidth]{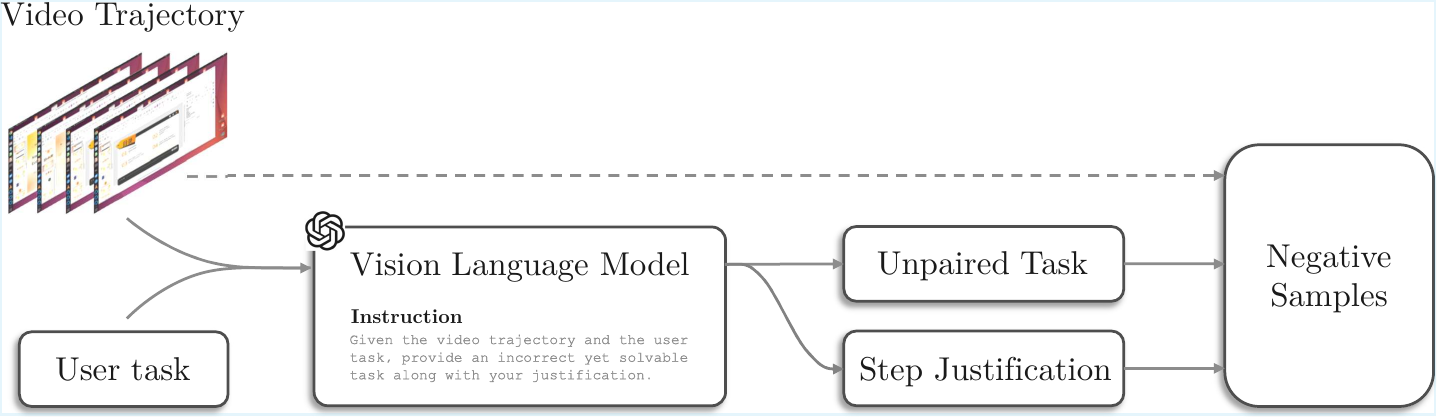}
  \caption{Illustration of how we synthesize negative samples via adversarial instruction translation. We use GPT-5.2 as the Vision Language Model.}
  \label{fig:neg_sample}
\end{figure}

AgentNet and ScaleCUA predominantly provide positive supervision because they are collected as successful demonstrations for training GUI agents. To obtain informative counterexamples for reward modeling, we introduce a simple yet effective synthetic-negative construction procedure, which we term \emph{adversarial translation}, inspired by back-translation~\cite{backtranslation}. As illustrated in \autoref{fig:neg_sample}, we start from a valid trajectory segment (a short sequence of UI states and actions) and prompt a vision--language model to generate an \emph{unpaired} task instruction that is plausible in the same interface context but does not match the demonstrated segment. During this translation process, the model is required to output both (i) a brief justification of why the segment violates the generated instruction and (ii) a reference step (i.e., the time index where the mismatch first becomes evident); we use these justifications as attribution labels for temporal grounding. This yields hard negatives that are visually similar yet semantically inconsistent with the intended goal. We then manually verify these synthetic pairs and retain only high-quality negatives; in our annotation audit, the resulting negatives achieve a 100\% human pass rate on a selected subset, providing reliable contrastive supervision for training our reward model.

\subsection{Spatiotemporal Token Pruning}
\begin{algorithm}[h]
  \caption{Training w/ Spatiotemporal Token Pruning}
  \label{alg:overall}
  \small
  \begin{algorithmic}[1]

    \Require Training dataset $\mathcal{D}_{\text{train}}$, vision--language model $\mathcal{M}$, STP threshold $\tau_s$, TTP threshold $\tau_t$, component threshold $\tau_{\text{large}}$

    \State Freeze vision encoder $\mathcal{V}$ and projector $\mathcal{P}$ \Comment{Only train LLM parameters}

    \For{$(X, V, Y) \sim \mathcal{D}_{\text{train}}$}
    \Comment{$X$: text tokens, $V$: video frames, $Y$: labels}

    \vspace{2mm}
    \State \textcolor{softblue}{\textbf{// Token Pruning}}
    \State $Z_V \gets \mathcal{V}(V)$
    \Comment{Encode video frames into patch tokens}
    \State $\mathbf{M}_s \gets \mathrm{STP}(Z_V, \tau_s, \tau_{\text{large}})$
    \Comment{\autoref{alg:str}}
    \State $\mathbf{M}_t \gets \mathrm{TTP}(Z_V, \tau_t)$
    \Comment{\autoref{alg:ttr}}
    \State $\mathbf{M} \gets \mathbf{M}_s \land \mathbf{M}_t$
    \Comment{Token kept only if both agree}

    \vspace{2mm}
    \State \textcolor{softblue}{\textbf{// Apply Masks to Video Tokens}}
    \State $\widetilde{Z}_V \gets \mathrm{Pack}(Z_V, \mathbf{M})$
    \Comment{Drop pruned tokens and re-pack sequence}
    \State $\widetilde{H}_V \gets \mathcal{P}(\widetilde{Z}_V)$
    \Comment{Project visual tokens to LLM input space}

    \vspace{2mm}
    \State \textcolor{softblue}{\textbf{// Update Model}}
    \State $\hat{Y} \gets \mathcal{M}_{\text{LLM}}(\widetilde{H}_V, X)$
    \State $\mathcal{L}_{\text{rm}} \gets \ell(\hat{Y}, Y)$
    \State Update trainable parameters with $\nabla \mathcal{L}_{\text{rm}}$

    \EndFor

  \end{algorithmic}
\end{algorithm}
Computer-use trajectory videos contain a large number of fine-grained elements at different scales, such as windows, icons, menus, cursors, and small text. Training reward models for computer use often benefits from high-resolution video inputs, where these subtle UI cues are clearly visible. However, the computational cost of processing high-resolution frames is substantial: naively placing all frames at full resolution into the model context quickly exhausts the token budget and leads to prohibitive memory usage, making it difficult to fit long interaction trajectories into a single forward-backward pass. To address this issue, we propose \emph{Spatiotemporal Token Pruning}, which can significantly reduce the effective context length by pruning redundant visual tokens in both space and time. As summarized in \autoref{alg:overall}, we combine Spatial Token Pruning (STP; \autoref{sec:str}) to discard spatially redundant background regions within each frame, and Temporal Token Pruning (TTP; \autoref{sec:ttr}) to suppress temporally repeated tokens across frames, enabling efficient training with high-resolution videos without sacrificing the transient UI evidence critical for decision making.

\subsubsection{Spatial Token Pruning}
\label{sec:str}
\begin{algorithm}[t]
  \caption{Spatial Token Pruning (STP)}
  \label{alg:str}
  \small
  \begin{algorithmic}[1]
    \Require Patch features $\mathbf{P}$, adjacency threshold $\tau_s$, large-component threshold $\tau_{\text{large}}$
    \State Reshape $\mathbf{P}\in\mathbb{R}^{N\times D}$ into per-frame grids $\mathbf{P}^{(t)}\in\mathbb{R}^{H'\times W'\times D}$, $t\in[T]$
    \For{$t \gets 1$ \textbf{to} $T$.}
    \State Compute neighbor distances $d_h^{(t)}(i,j)$ and $d_v^{(t)}(i,j)$ from $\mathbf{P}^{(t)}$.
    \State Define adjacency indicator $e^{(t)}_{(i,j)\leftrightarrow(i',j')}$ using $\tau_s$.
    \State Build graph $\mathcal{G}^{(t)}=(V^{(t)},E^{(t)})$ from $e_{(i,j)\leftrightarrow(i',j')}$.
    \State Obtain connected components $\mathcal{C}^{(t)}=\{C_k^{(t)}\}$.
    \State Mark large components $\mathcal{R}^{(t)}\subseteq\mathcal{C}^{(t)}$ using $\tau_{\text{large}}$.
    \State Construct spatial mask $\mathbf{M}_s^{(t)}\in\{0,1\}^{H'\times W'}$ from $\mathcal{R}^{(t)}$.
    \EndFor
    \State \Return $\{\mathbf{M}_s^{(t)}\}_{t=1}^{T}$
  \end{algorithmic}
\end{algorithm}

Let $\mathbf{P} \in \mathbb{R}^{N \times D}$ denote the patch-feature matrix produced by the frozen vision encoder, where $N=T \times H' \times W'$ is the total number of spatiotemporal patches across $T$ frames on an $H' \times W'$ grid, and $D$ is the feature dimension. Inspired by ShowUI~\cite{showui}, we construct a per-frame UI-connected graph $\mathcal{G}^{(t)}=(V^{(t)},E^{(t)})$, where each node in $V^{(t)}$ corresponds to a patch and edges in $E^{(t)}$ connect spatial neighbors with similar features. The resulting connected components are denoted by $\mathcal{C}^{(t)}=\{C_1^{(t)},\ldots,C_{K_t}^{(t)}\}$. We use $\tau_s$ as the similarity threshold for establishing edges, and $\tau_{\text{large}}$ as the size threshold for identifying components to be removed. For each frame $t$, STP outputs a binary mask $\mathbf{M}^{(t)}_s \in \{0,1\}^{H' \times W'}$ indicating which patches are retained.

For a single frame $t$, we reshape patch features into a grid $\mathbf{P}^{(t)} \in \mathbb{R}^{H' \times W' \times D}$, where $H'=H/m$ and $W'=W/m$ for merge size $m$. We first compute local feature differences between neighboring patches. The horizontal difference is
\begin{equation}
  d^{(t)}_{h}(i, j) = \left\| \mathbf{P}^{(t)}_{i,j} - \mathbf{P}^{(t)}_{i,j+1} \right\|_2, \quad 1 \le i \le H',\; 1 \le j < W',
\end{equation}
while the vertical difference is
\begin{equation}
  d^{(t)}_{v}(i, j) = \left\| \mathbf{P}^{(t)}_{i,j} - \mathbf{P}^{(t)}_{i+1,j} \right\|_2, \quad 1 \le i < H',\; 1 \le j \le W'.
\end{equation}
Let $\mathcal{N}(i,j)$ be the 4-neighborhood of location $(i,j)$. Two patches are connected if they are neighbors and their feature distance is below $\tau_s$:
\begin{equation}
  e^{(t)}_{(i,j) \leftrightarrow (i',j')} =
  \mathds{1}\!\left[(i',j')\in\mathcal{N}(i,j)\;\land\;\left\| \mathbf{P}^{(t)}_{i,j} - \mathbf{P}^{(t)}_{i',j'} \right\|_2 < \tau_s\right].
\end{equation}

We then apply Union-Find on $\mathcal{G}^{(t)}$ to obtain connected components and select large components for pruning:
\begin{equation}
  \mathcal{C}^{(t)} = \text{Union-Find}\!\left(V^{(t)}, E^{(t)}\right), \quad
  \mathcal{R}^{(t)} = \left\{ C \in \mathcal{C}^{(t)} : |C| > \tau_{\text{large}} \right\}.
\end{equation}

Let $C^{(t)}(i,j)$ denote the component containing patch $(i,j)$. The spatial mask is
\begin{equation}
  \mathbf{M}^{(t)}_s(i,j)=
  \begin{cases}
    0 & \text{if } C^{(t)}(i,j) \in \mathcal{R}^{(t)}, \\
    1 & \text{otherwise}.
  \end{cases}
\end{equation}
% Intuitively, large components often correspond to visually homogeneous regions (e.g., wallpapers or window backgrounds) that consume a substantial fraction of the token budget while providing limited UI-discriminative evidence, whereas smaller components tend to capture localized elements such as icons, buttons, cursors, and text blocks. 
Applying $\mathbf{M}^{(t)}_s$ filters the original token grid and packs the remaining tokens into a shorter sequence, which is then fed to the vision-language model. This parameter-free procedure preserves local UI structure and effectively reduces redundant spatial tokens while retaining fine-grained cues.

\begin{algorithm}[t]
  \caption{Temporal Token Pruning (TTP)}
  \label{alg:ttr}
  \small
  \begin{algorithmic}[1]
    \Require Video tokens $\mathbf{V}\in\mathbb{R}^{T\times N\times D}$, threshold $\tau_t$, similarity $\text{sim}_{\text{cos}}(\cdot,\cdot)$

    \For{$i=0$ \textbf{to} $N-1$}
    \State $\mathbf{v}^{(\text{ref})}_i \gets \mathbf{v}^{(0)}_i$ \Comment{Frame $0$ is the initial reference frame}
    \State $\mathbf{M}_t(0,i) \gets 1$ \Comment{Always keep reference frame}
    \For{$t=1$ \textbf{to} $T-1$}
    \If{$\text{sim}_{\text{cos}}(\mathbf{v}^{(\text{ref})}_i,\mathbf{v}^{(t)}_i) > \tau_t$}
    \State $\mathbf{M}_t(t,i) \gets 0$ \Comment{Similar to reference, remove}
    \Else
    \State $\mathbf{M}_t(t,i) \gets 1$ \Comment{Different from reference, keep}
    \State $\mathbf{v}^{(\text{ref})}_i \gets \mathbf{v}^{(t)}_i$ \Comment{Update reference}
    \EndIf
    \EndFor
    \EndFor
    \State \Return $\mathbf{M}_t$
  \end{algorithmic}
\end{algorithm}
\subsubsection{Temporal Token Pruning}
\label{sec:ttr}

In UI-dense videos, spatial redundancy alone is often insufficient: many frames share the same layout (menus, sidebars, toolbars, and background windows), while task-relevant evidence is concentrated in small, transient changes (e.g., a highlighted button, a newly opened dropdown, a scrolling offset, or a cursor-driven text update). In such cases, STP may still retain many tokens per frame because the UI contains high-contrast elements everywhere, even though most of them remain unchanged across adjacent frames. Moreover, STP can be unreliable in the presence of visually significant yet semantically irrelevant content (e.g., image-dense webpages). Therefore, we further propose temporal token pruning (TTP), which complements STP by removing temporally repeated tokens and reallocating the token budget to regions that actually evolve over time.

We denote the video token tensor by $\mathbf{V}\in\mathbb{R}^{T\times N\times D}$, where $T$ is the number of frames, $N$ is the number of (merged) spatial tokens per frame, and $D$ is the token dimension. Let $\mathbf{v}^{(t)}_i\in\mathbb{R}^D$ be the token at spatial location $i$ in frame $t$. Analogous to STP, our goal is to identify and remove tokens that contribute little new information under the UI-centric computer-using setting. While STP suppresses spatially homogeneous regions within each frame, TTP targets redundancy across time by filtering tokens whose appearance remains nearly unchanged over consecutive frames.

Specifically, for each spatial location $i$, we maintain a reference token $\mathbf{v}^{(\text{ref})}_i$ initialized from the first frame and compare subsequent tokens against this reference using cosine similarity $\text{sim}_{\text{cos}}(\cdot,\cdot)$. We set the temporal mask as
\begin{equation}
  \mathbf{M}_t(t,i)=\mathds{1}\!\left[\text{sim}_{\text{cos}}\!\left(\mathbf{v}^{(\text{ref})}_i, \mathbf{v}^{(t)}_i\right) \le \tau_t\right], \quad t\in\{1,\ldots,T-1\}.
\end{equation}
We always keep the first-frame token by setting $\mathbf{M}_t(0,i)=1$. Equivalently, a token is pruned when $\text{sim}_{\text{cos}}(\mathbf{v}^{(\text{ref})}_i, \mathbf{v}^{(t)}_i) > \tau_t$. The reference update follows
\begin{equation}
  \mathbf{v}^{(\text{ref})}_i \leftarrow
  \begin{cases}
    \mathbf{v}^{(t)}_i & \text{if } \mathbf{M}_t(t,i)=1, \\
    \mathbf{v}^{(\text{ref})}_i & \text{otherwise}.
  \end{cases}
\end{equation}
where $\tau_t$ is the similarity threshold. This update rule ensures that each location is compared against its most recent ``distinct'' state, effectively capturing step-wise UI transitions (cursor moves, menu expansion, scrolling, and window switching) while ignoring static backgrounds. Finally, \autoref{alg:ttr} constructs a binary temporal mask $\mathbf{M}_t\in\{0,1\}^{T\times N}$ which can be applied to the token sequence to drop temporally redundant tokens. 
% Similar to STP, TTP is parameter-free and only depends on $\tau_t$ and the similarity function; in practice it reduces the effective token budget for long videos while preserving the transient UI evidence that is most relevant for downstream decision making.

\section{Experiments}

\subsection{Implementation Details}

Our approach introduces three key hyperparameters controlling token filtering: the spatial threshold $\tau_s$ for STP, the temporal similarity threshold $\tau_t$ for TTP, and a ``large-component'' threshold $\tau_{\text{large}}$ used to guard against over-pruning when abrupt UI transitions occur. Intuitively, decreasing $\tau_s$ makes STP more aggressive by removing a larger portion of spatially low-saliency tokens, while increasing $\tau_s$ retains more within-frame context at the cost of a higher token budget. Similarly, increasing $\tau_t$ makes TTP stricter (i.e., only near-identical tokens are removed), whereas lowering $\tau_t$ prunes more temporal repeats but risks discarding subtle yet task-relevant UI updates. We set $\tau_s=0.3$, $\tau_t=0.9999$, and $\tau_{\text{large}}=40$ for all experiments. We instantiate our video reward model from \texttt{Qwen3-VL-4B-Instruct} and \texttt{Qwen3-VL-8B-Instruct}, and fine-tune with a learning rate of $5\times 10^{-6}$ using a cosine decay schedule. All training runs are performed on 8$\times$NVIDIA A100 GPUs with 80\,GB memory each. Our training is based on a modified LLaMA-Factory~\cite{zheng2024llamafactory} framework.

\subsection{Evaluation Protocol}

\subsubsection{\testset}
We build our evaluation benchmark, \testset, from a held-out split of \dataset. Each instance pairs a \textit{user instruction} with a \textit{video trajectory}. We evaluate two settings: (i) binary judgment (\texttt{correct} vs.\ \texttt{incorrect}), and (ii) \textit{attribution judgment}, which additionally requires a short time range indicating where the first error occurs. The initial pool contains 800 tasks, evenly split across Ubuntu (Agent), Ubuntu (Human), Mac/Win, and Android. For Ubuntu (Agent), we sample 10 tasks from OSWorld and collect 200 distinct solutions to capture trajectory diversity under shared goals; for Ubuntu (Human), Mac/Win, and Android, we separate 200 unique task--solution pairs per category from \dataset to broaden task coverage. We also annotate 200 instances with first-deviation time ranges for fine-grained temporal localization evaluation. For consistent visual input, all videos are rendered at 720p, and models receive up to 100 frames at 1 FPS (uniformly sampled when longer). After removing unsolvable tasks, \testset contains 789 instances, with an approximately balanced class ratio (49.94\% positive vs.\ 50.06\% negative).

\subsubsection{Evaluation metrics}
We report standard classification metrics, including accuracy, precision, and recall, to measure how well the reward model distinguishes positive from negative trajectories. In addition, to evaluate temporal grounding quality, \ie, whether the model localizes the critical time span responsible for failure, we compute a temporal intersection-over-union (tIoU) between the model-predicted interval $\hat{\mathcal{I}}=[\hat{t}_s,\hat{t}_e]$ and the ground-truth interval ${\mathcal{I}}=[t_s,t_e]$:
\begin{equation}
  \mathrm{tIoU}(\hat{\mathcal{I}}, \mathcal{I}) = \frac{|\hat{\mathcal{I}} \cap \mathcal{I}|}{|\hat{\mathcal{I}} \cup \mathcal{I}|}
  = \frac{\max\!\left(0,\, \min(\hat{t}_e,t_e) - \max(\hat{t}_s,t_s)\right)}{\max(\hat{t}_e,t_e) - \min(\hat{t}_s,t_s)}.
\end{equation}
This metric rewards accurate localization and penalizes overly broad predictions.
% , providing a complementary view to global correctness metrics in UI-centric, step-based videos.

\subsection{Main Result}

We report the main quantitative comparison on \testset in \autoref{tab:main_res} and \autoref{fig:tiou_comp}. Overall, \model consistently outperforms both proprietary and open-sourced baselines, indicating that explicitly training a video reward model on computer-using trajectories yields stronger preference modeling than directly prompting general-purpose VLMs.
\begin{table*}[t]
  \centering
  \caption{Detailed performance on Agent/Human trajectory evaluation on \testset.
  We report Accuracy (Acc), Precision (Prec), and Recall (Rec). \underline{Underline} indicates the second-best result, and \textbf{bold} indicates the best result. \model~8B achieves the best overall balance, outperforming open-source baselines and matching or exceeding strong proprietary models on most settings.}
  \label{tab:main_res}
  \scriptsize
  \setlength{\tabcolsep}{2.8pt}
  \resizebox{\textwidth}{!}{%
    \begin{tabular}{lccc|ccc|ccc|ccc|ccc}
      \toprule
      \multirow{3}{*}{\textbf{Model}}
      & \multicolumn{3}{c}{\textbf{Ubuntu (Agent)}}
      & \multicolumn{3}{c}{\textbf{Ubuntu (Human)}}
      & \multicolumn{3}{c}{\textbf{Mac/Win}}
      & \multicolumn{3}{c}{\textbf{Android}}
      & \multicolumn{3}{c}{\textbf{Overall}} \\
      \cmidrule(lr){2-4}\cmidrule(lr){5-7}\cmidrule(lr){8-10}\cmidrule(lr){11-13}\cmidrule(lr){14-16}
      & \textbf{Acc.} & \textbf{Prec.} & \textbf{Rec.}
      & \textbf{Acc.} & \textbf{Prec.} & \textbf{Rec.}
      & \textbf{Acc.} & \textbf{Prec.} & \textbf{Rec.}
      & \textbf{Acc.} & \textbf{Prec.} & \textbf{Rec.}
      & \textbf{Acc.} & \textbf{Prec.} & \textbf{Rec.} \\
      \midrule
      \multicolumn{16}{c}{\textit{Proprietary Models}} \\
      \midrule
      Gemini 2.5 Pro~\cite{gemini25} & \underline{84.6} & \underline{85.6} & 82.8 & 64.5 & 82.2 & 37.0 & 69.0 & 83.9 & 47.0 & 74.0 & \underline{81.6} & 62.0 & 72.8 & \underline{83.5} & 56.7 \\
      Gemini 3 Pro~\cite{gemini3}   & 80.4 & 75.0 & \textbf{90.0} & 71.2 & 71.7 & 71.0 & 75.6 & 75.3 & 76.8 & 73.5 & 74.7 & 71.0 & 75.1 & 74.2 & 76.7 \\
      GPT-5.2~\cite{gpt52}        & 82.5 & 85.1 & 78.7 & 74.0 & \textbf{84.3} & 59.0 & 74.5 & \textbf{88.9} & 68.7 & 74.5 & 75.5 & 75.8 & 75.0 & 82.7 & 66.5 \\
      Seed-2.0 Pro~\cite{seed2}    & \textbf{85.1} & 85.1 & \underline{85.1} & 77.2 & 81.7 & 69.1 & 81.0 & 86.1 & 74.0 & \underline{78.0} & \textbf{82.6} & 71.0 & \underline{80.3} & \textbf{83.9} & 74.7 \\
      \midrule
      \multicolumn{16}{c}{\textit{Open-sourced Models}} \\
      \midrule
      LLaVA-Next-Video 7B~\cite{llava-next} & 50.8 & 51.6 & 17.0 & 48.5 & 46.7 & 21.0 & 49.5 & 48.8 & 21.0 & 49.0 & 47.5 & 19.0 & 13.6 & 48.4 & 19.4 \\
      InternVL-3.5 8B~\cite{intern35}  & 64.6 & 64.8 & 55.9 & 52.0 & 52.2 & 48.0 & 59.5 & 58.6 & 65.0 & 57.5 & 59.3 & 48.0 & 56.6 & 58.9 & 55.9 \\
      Qwen2.5-VL 7B~\cite{qwen25}  & 64.6 & 65.5 & 60.6 & 63.0 & 65.1 & 56.0 & 68.0 & 74.3 & 55.0 & 63.5 & 62.7 & 66.0 & 64.5 & 66.6 & 59.2 \\
      Qwen3-VL 4B~\cite{qwen3}    & 76.2 & 72.9 & 82.9 & 72.5 & 76.5 & 65.0 & 74.5 & 74.3 & 75.0 & 68.5 & 70.8 & 63.0 & 72.9 & 73.7 & 71.3 \\
      Qwen3-VL 8B~\cite{qwen3}    & 71.4 & 74.4 & 64.9 & 69.0 & 86.5 & 45.0 & 66.5 & 81.1 & 43.0 & 64.5 & 71.6 & 48.0 & 67.7 & 77.6 & 49.9 \\
      \midrule
      \model 4B       & 77.8 & 83.3 & 69.1 & \underline{81.0} & 82.9 & \underline{78.0} & \textbf{90.0} & \underline{87.7} & \underline{93.0} & 72.5 & 66.4 & \underline{91.0} & 80.1 & 79.2 & \underline{82.5} \\
      \model 8B       & 82.5 & \textbf{85.9} & 77.7 & \textbf{84.0} & \underline{84.0} & \textbf{84.0} & \underline{89.0} & 85.5 & \textbf{94.0} & \textbf{83.5} & 77.2 & \textbf{95.0} & \textbf{84.7} & 82.9 & \textbf{87.7} \\
      \bottomrule
    \end{tabular}%
  }
\end{table*}

\subsubsection{Performance comparison}
Consistent with \autoref{tab:main_res}, \model~8B ranks first overall with \textbf{84.7} accuracy, \textbf{82.9} precision, and \textbf{87.7} recall, surpassing the strongest proprietary baselines Seed-2.0 Pro (80.3/83.9/74.7) and GPT-5.2 (75.0/82.7/66.5). It also outperforms the best open-weight models by +17.1 accuracy points over Qwen3-VL~8B and +28.2 over InternVL-3.5~8B, while substantially improving recall (87.7 vs.\ 49.9 and 55.9). Gains are consistent across settings: Mac/Win reaches \underline{89.0} accuracy with \textbf{94.0} recall, Android reaches \textbf{83.5} accuracy with \textbf{95.0} recall, and Ubuntu reaches 82.5/77.7 (Agent) and \textbf{84.0}/\textbf{84.0} (Human) for accuracy/recall, all leading their baselines. Scaling from 4B to 8B further brings +4.6 overall accuracy (84.7 vs.\ 80.1) and +5.2 recall (87.7 vs.\ 82.5), with the largest gains on Android (+11.0) and Ubuntu (Agent, +4.8).

\subsubsection{Attribution comparison}
\begin{wrapfigure}{r}{0.52\textwidth}
  \centering
  \includegraphics[width=0.5\textwidth]{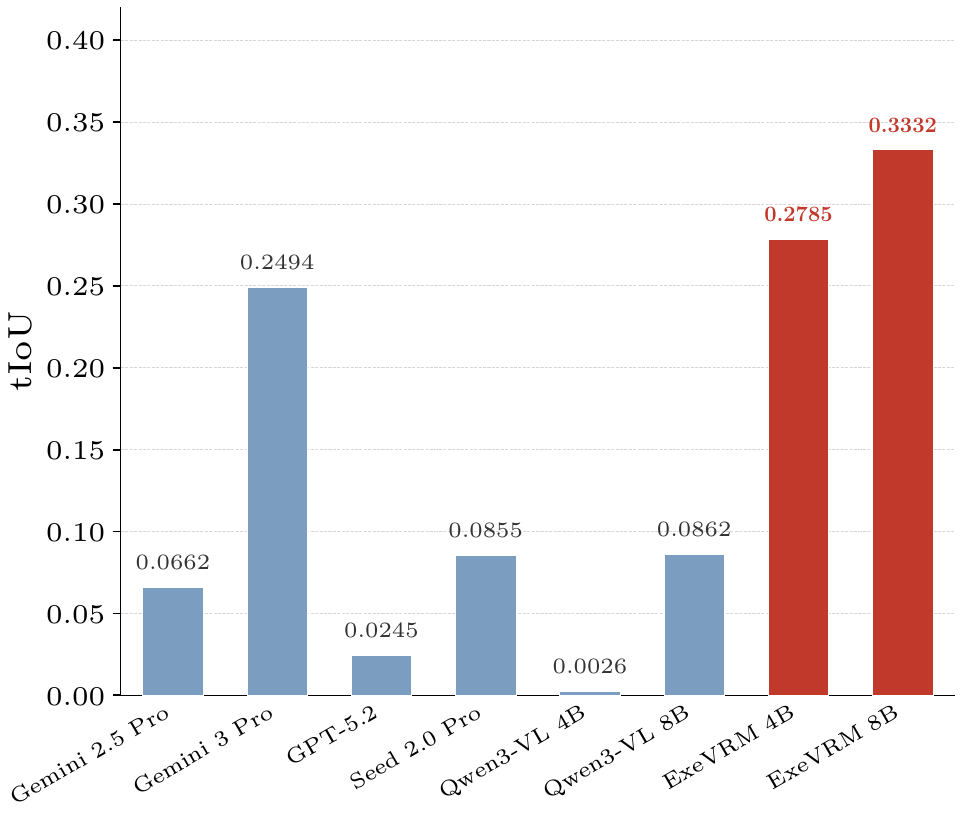}
  \caption{Comparison of temporal IoU (tIoU) scores across models on \testset.}
  \vspace{-10pt}
  \label{fig:tiou_comp}
\end{wrapfigure}

Beyond overall accuracy, we analyze attribution quality: \emph{how well a model can localize the decisive temporal window that determines whether a trajectory succeeds or fails.} We compute temporal IoU (tIoU) between each model's predicted salient segment and human-annotated evidence spans on \testset.
As shown in \autoref{fig:tiou_comp}, \model attains consistently higher tIoU than all baselines, indicating more precise and less noisy credit assignment over time. This suggests that explicitly training a reward model on computer-using trajectories not only improves preference prediction, but also yields explanations that better align with human judgments about when the outcome is decided. In practice, stronger attribution makes downstream agent development easier: it highlights the exact interaction steps that cause failures (e.g., an incorrect click or a missed UI state change), enabling faster debugging and more targeted data collection.

\begin{table}[h]
\centering
\setlength{\tabcolsep}{5pt}
\caption{Performance comparison across different evaluation strategies. Here Tail (T) means the last screenshot and Head (H) means the first screenshot. Compact denotes a compact list of action history, while Details denotes a structured action history. Oneshot means there is an example in the prompt as a reference. }
\label{tab:eval_strategies}
\resizebox{\textwidth}{!}{%
\begin{tabular}{@{}llccccccccc@{}}
\toprule
\multirow{2}{*}{\textbf{Model}} & \multirow{2}{*}{\textbf{Method}} & \multirow{2}{*}{\textbf{Res.}} & \multirow{2}{*}{\textbf{Visual}} & \multirow{2}{*}{\textbf{Info.}} & \multicolumn{3}{c}{\textbf{Ubuntu(Human)}} & \multicolumn{3}{c}{\textbf{Mac/Win}} \\
\cmidrule(lr){6-8} \cmidrule(lr){9-11}
& & & & & Acc. & Prec. & Rec. & Acc. & Prec. & Rec. \\
\midrule
\multirow{5}{*}{Qwen3-VL 4B} 
 & AER~\cite{AER} & 720p & Tail & Compact & 58.5 & 53.3 & 70.3 & 69.0 & 84.0 & 72.9 \\
 & Simplified Judge~\cite{simplifiedjudge} & 720p &H+T & Details & 57.5 & 52.1 & \textbf{81.3} & 70.8 & 87.1 & 71.5 \\
 & SE-WSM~\cite{sun2025seagentselfevolvingcomputeruse} & 360p &Full & None & 50.5 & 44.6 & 36.3 & 46.0 & 84.1 & 35.1\\
 & ZeroGUI~\cite{yang2025zeroguiautomatingonlinegui} & 360p &Full & Oneshot & 48.0 & 42.5 & 40.7 & 51.5 & 81.4 & 46.4 \\
 & \model 4B (Ours) & 720p & Video & None & \textbf{81.0} & \textbf{82.9} & 78.0 & \textbf{90.0} & \textbf{87.7} & \textbf{93.0} \\
\midrule
\multirow{5}{*}{Qwen3-VL 8B} 
 & AER~\cite{AER} & 720p & Tail & Compact & 55.0 & 50.5 & 61.5 & 68.0 & 80.4 & 76.2 \\
 & Simplified Judge~\cite{simplifiedjudge} & 720p & H+T & Details & 58.5 & 53.5 & 68.1 & 78.2 & \textbf{89.6} & 80.1 \\
 & SE-WSM~\cite{sun2025seagentselfevolvingcomputeruse} & 360p & Full & None & 48.5 & 43.2 & 41.8 & 51.0 & 81.5 & 46.4 \\
 & ZeroGUI~\cite{yang2025zeroguiautomatingonlinegui} & 360p & Full & Oneshot & 50.0 & 46.7 & 61.5 & 63.5 & 81.5 & 66.9 \\
 & \model 8B (Ours) & 720p & Video & None & \textbf{84.0} & \textbf{84.0} & \textbf{84.0} & \textbf{89.0} & 85.5 & \textbf{94.0} \\
\bottomrule
\end{tabular}
}
\vspace{-10pt}
\end{table}
\section{Discussion \& Ablation Studies}

\subsubsection{Findings 1: Dense video context outperforms sparse snapshots}
\label{subsubsec:findings1}
Previous work on task-completion assessment in CUA largely relies on screenshots rather than execution video. In particular, AER~\cite{AER} evaluates only the final-state screenshot, while Simplified Judge~\cite{simplifiedjudge} add initial screenshot into considerations. However, these sparse-observation methods underperform (\autoref{tab:eval_strategies}), indicating that ignoring causal transitions throughout the interaction makes accurate completion judgment difficult. SE-WSM~\cite{sun2025seagentselfevolvingcomputeruse} and ZeroGUI~\cite{yang2025zeroguiautomatingonlinegui} instead feed every key frame to the model, which leads to OOM on a single A100 (80GB) without token pruning. If we reduce the resolution to 360p to meet the constraint, performance can even fall below AER~\cite{AER} that only judges final state. This finding highlights the necessity of both video-based assessment and token pruning.

\subsubsection{Findings 2: Higher resolution brings more benefit for reward modeling}
\begin{table}[t]
  \center
  \caption{Effect of input resolution on reward-model performance. Increasing the input from 360p to 720p with spatiotemporal token pruning (STP\&TTP) consistently improves accuracy and recall for both Qwen3-VL 4B and 8B.}
  \label{tab:resolution-abla}
  \resizebox{0.77\textwidth}{!}{%
    \setlength{\tabcolsep}{5pt}
    \begin{tabular}{@{}llccc@{}}
      \toprule
      \textbf{Model} & \textbf{Resolution} & \textbf{Accuracy} & \textbf{Precision} & \textbf{Recall} \\ \midrule
      \multirow{2}{*}{Qwen3-VL 4B} & 360p                    & 79.3 & \textbf{80.6 }& 77.8 \\
      & 720p (w/STP \& TTP)  & \textbf{80.1} & 79.2 & \textbf{82.5} \\ \midrule
      \multirow{2}{*}{Qwen3-VL 8B} & 360p                    & 81.5      & 82.5      &  80.5     \\
      & 720p (w/STP \& TTP) & \textbf{84.7} & \textbf{82.9} & \textbf{87.7}      \\ \bottomrule
    \end{tabular}%
  }
\end{table}
As shown in \autoref{tab:resolution-abla}, moving from 360p to 720p yields clear gains in completion judgment quality, especially in recall. For Qwen3-VL 4B, accuracy improves from 79.3 to 80.1 and recall rises from 77.8 to 82.5 (+4.7), with a modest precision drop (80.6$\rightarrow$79.2). For Qwen3-VL 8B, all three metrics improve: accuracy increases from 81.5 to 84.7 (+3.2), precision from 82.5 to 82.9, and recall from 80.5 to 87.7 (+7.2). These results indicate that higher-resolution inputs preserve fine-grained GUI cues (e.g., small text edits and localized state changes) that are frequently decisive for reward prediction, while STP/TTP keeps 720p training and inference tractable.
\subsubsection{Findings 3: Asymmetric Effects of Spatial and Temporal Pruning}

\begin{table}[t]
  \center
  \caption{Ablation of spatiotemporal token pruning on Qwen3-VL 4B. STP denotes spatial token pruning and TTP denotes temporal token pruning. TTP contributes stronger gains than STP when used alone, while the full STP+TTP configuration remains competitive and achieves the highest recall.}
  \label{tab:stp-ttp-able}
  \resizebox{0.8\textwidth}{!}{%
    \setlength{\tabcolsep}{5pt}
    \begin{tabular}{@{}lccccc@{}}
      \toprule
      \textbf{Model} & \textbf{w/STP} & \textbf{w/TTP} & \textbf{Accuracy} & \textbf{Precision} & \textbf{Recall} \\ \midrule
      \multirow{3}{*}{Qwen3-VL 4B} & \checkmark &   & 77.6 & 81.7 & 72.6 \\
      &   & \checkmark & 80.3      & 81.3      & 79.3      \\
      & \checkmark & \checkmark & 80.1      & 79.2      & 82.5      \\ \bottomrule
    \end{tabular}%
  }
\end{table}

\autoref{tab:stp-ttp-able} reveals an asymmetric contribution of the two pruning modules. Applying STP alone yields lower accuracy and recall (77.9/72.6), whereas TTP alone provides the strongest overall balance (80.3/79.3). We conjecture this trend follows from the nature of GUI trajectories: within-frame saliency is often dominated by visually rich yet task-irrelevant regions (e.g., dense menus, icons, and textured webpage content), so spatial pruning may occasionally discard fine-grained cues that are weak visually but decisive for completion judgment. By contrast, reward prediction is primarily driven by inter-frame state transitions, which are directly targeted by TTP through suppressing temporally repeated tokens and preserving changing evidence. Notably, enabling STP and TTP jointly causes only a marginal accuracy change relative to TTP-only (80.1 vs. 80.3) while improving recall to 82.5, suggesting that the full configuration remains robust and mainly induces a precision--recall trade-off rather than a substantial performance drop.

\subsubsection{Findings 4: Spatiotemporal pruning improves training efficiency}

\begin{figure}[t]
  \centering
  \begin{minipage}[t]{0.49\linewidth}
    \centering
    \includegraphics[width=\linewidth]{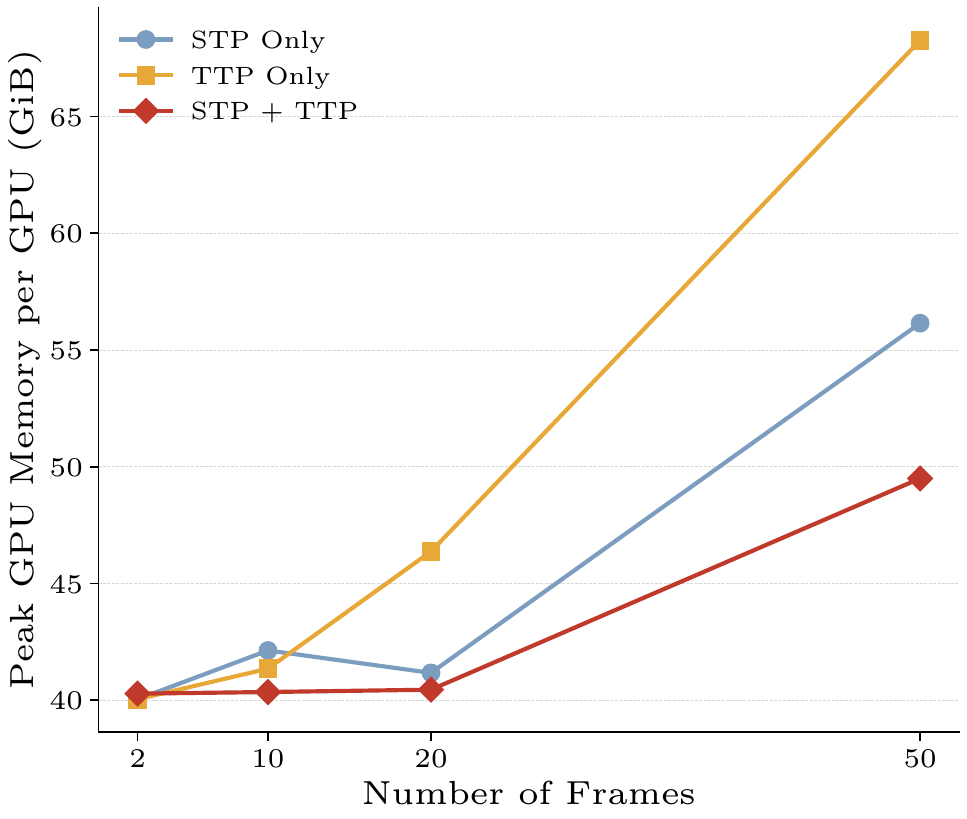}
  \end{minipage}\hfill
  \begin{minipage}[t]{0.49\linewidth}
    \centering
    \includegraphics[width=\linewidth]{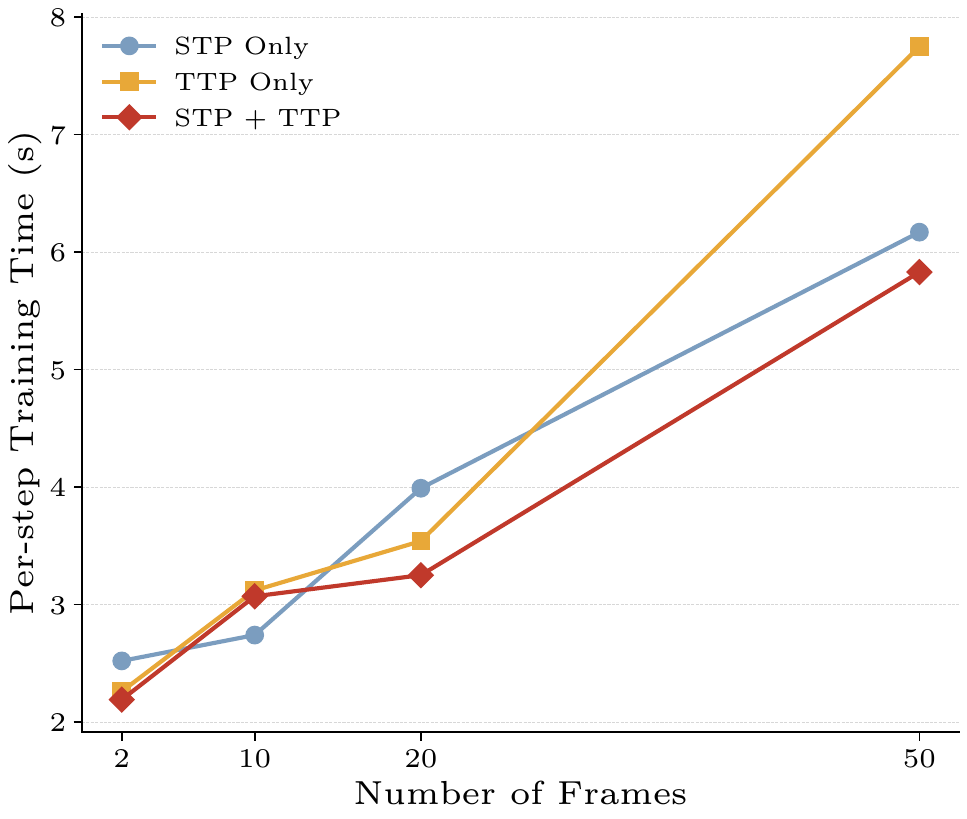}
  \end{minipage}
  \caption{Efficiency analysis. Left: memory usage. Right: runtime.}
\label{fig:efficiency}
\end{figure}

As shown in \autoref{fig:efficiency}, both peak GPU memory and per-step training time increase with the number of input frames for all variants, but the growth rates differ substantially. With 50 frames, STP-only reaches about 56\,GiB memory and 6.2\,s/step, while TTP-only rises further to about 68\,GiB and 7.8\,s/step. In contrast, combining STP and TTP keeps the footprint much lower ($\sim$49.5\,GiB and 5.8\,s/step), and remains close to the 40\,GiB regime up to 20 frames. This trend is consistent with our method design: spatial pruning removes large homogeneous regions in each frame, whereas temporal pruning suppresses repeated tokens across adjacent frames. Jointly applying both reduces redundancy along both spatial and temporal dimensions, so the benefit becomes more pronounced as trajectories get longer. These efficiency gains are what make 720p long-horizon reward-model training practical under our hardware setup (8$\times$A100-80GB), without resorting to aggressive resolution downsampling. Additional analyses and visualizations of STP and TTP are provided in the supplementary materials. %Appendix~\ref{apdx:stp-ttp-visualize}.

\section{Conclusion}
In this work, we present a video-execution paradigm for reward modeling of computer-using agents (CUA). We introduce \dataset, a 53k--scale corpus of instruction--video--reward triplets, and propose adversarial instruction translation to synthesize hard negative pairs with step-level mismatch signals while producing justification-based attribution labels for temporal grounding. To make long-horizon, high-resolution execution videos trainable, we further develop spatiotemporal token pruning (STP+TTP), which reduces redundant spatial and temporal tokens while preserving decisive UI evidence. Extensive experiments on \testset show that our \model~8B achieves 84.7\% accuracy and 87.7\% recall, outperforming strong proprietary and open-weight baselines across Ubuntu, Mac/Win, and Android settings, with more accurate temporal attribution. Efficiency analysis further demonstrates that combining spatial and temporal pruning yields better memory--latency trade-offs as frame length grows. 

% Overall, our findings support video-execution reward modeling as a scalable, model-agnostic evaluator for CUA in realistic long-horizon environments.

\bibliographystyle{splncs04}
\bibliography{main}

\clearpage
\appendix
\section{Implications for Future Work}
Although \model shows strong overall performance on \testset, several limitations remain and motivate future work.

First, our current formulation is less effective on very long-horizon Ubuntu agent trajectories that contain substantial \emph{in-trajectory exploration}. In these cases, an agent may execute several locally unsuccessful attempts before eventually completing the task (e.g., repeatedly clicking nearby spreadsheet cells before selecting the correct target). Because our reward modeling setup is trained to judge success from compact execution video evidence under binary supervision, such trial-and-error patterns can be misinterpreted as failure-like behavior even when the final outcome is correct. More broadly, this reflects a gap between \emph{outcome-level} reward modeling and \emph{process-level} credit assignment for long trajectories.

Second, while STP+TTP significantly improves memory and runtime efficiency, long-horizon video reasoning remains expensive for ``think-in-the-image'' style multimodal models. The main bottleneck is token budget: increasing frame count, resolution, and reasoning depth simultaneously can quickly trigger out-of-memory issues and unstable training throughput. For this reason, our current study focuses on supervised fine-tuning rather than large-scale reinforcement learning with explicit long-form reasoning rollout. This is a practical design choice under current infrastructure constraints, not a claim that SFT is the optimal training paradigm for all settings.

A promising direction is to bridge these two limitations with process-aware supervision. One option is to decompose a long trajectory into intent-consistent sub-trajectories, for example by using video captioning or state-transition summarization to re-ground local user intent over time. This could convert our framework from a pure outcome judge into a process reward model (PRM) that evaluates intermediate progress and recovery behaviors. However, building such a benchmark requires fine-grained human annotations of step-level correctness and recovery phases, which we do not yet have at sufficient scale. Future work will study efficient annotation protocols, budget-constrained reasoning training, and RL-style optimization under strict thinking budgets so that long-horizon reasoning quality can improve without sacrificing tractability.

\section{Method list for OSWorld trajectory rollout}
To build a reward dataset with broad behavioral coverage, we intentionally roll out a diverse set of CUA systems on OSWorld, spanning both proprietary frontier models and open-weight research models. This design is important for our setting because \model is trained from \emph{execution video} only (without relying on internal thoughts or tool traces). If trajectories came from a narrow policy family, the reward model could overfit to agent-specific interaction habits rather than true task completion signals. We therefore prioritize diversity along multiple axes, including model family, parameter scale, alignment method, agent wrapper, and backend reasoning style.

Concretely, the evaluated systems include: UI-TARS, UI-TARS (72B, DPO), and Omnitars~\cite{uitars}; UI-TARS 2~\cite{uitars2}; Agent S2 with Gemini and o3 backends~\cite{agents2}; AGI-0~\cite{agi0}; AutoGLM and AutoGLM-V~\cite{autoglm}; Claude 3.7/4/4.5 Sonnet~\cite{claude37,claude4,claude45}; Doubao 1.5 thinking vision pro~\cite{doubao15}; Gemini 2.5~\cite{gemini25}; Jedi 7B with GPT 4o and o3~\cite{jedi}; kimi-vl-a3b~\cite{kimi-vl}; Agentic Lybic~\cite{agentic-lybic}; Mano~\cite{Mano}; MobileAgent V3~\cite{mobile-agent-v3}; o3~\cite{o3-o4}; GTA-1 (o3)~\cite{gta-1}; OpenCUA variants (32B, 7B, A3B, and Qwen2 7B)~\cite{opencua}; Qwen2.5-VL (32B/72B)~\cite{qwen25}; Tianxi Action 7B~\cite{tianxi-action}; and UIPath (GPT-5)~\cite{ilie2025uipath}.

Across this pool, agents exhibit substantial differences in action efficiency, recovery behavior, and error modes, producing both successful and near-miss trajectories under shared tasks. This heterogeneity is valuable for \dataset and complements our adversarial instruction translation pipeline: together they provide harder negatives, richer temporal divergence patterns, and more robust supervision for learning fine-grained success/failure signals. As a result, the reward model is better positioned to generalize across unseen CUA policies and platform-specific interaction styles.

\begin{figure}[h]
    \centering
    \includegraphics[width=\linewidth]{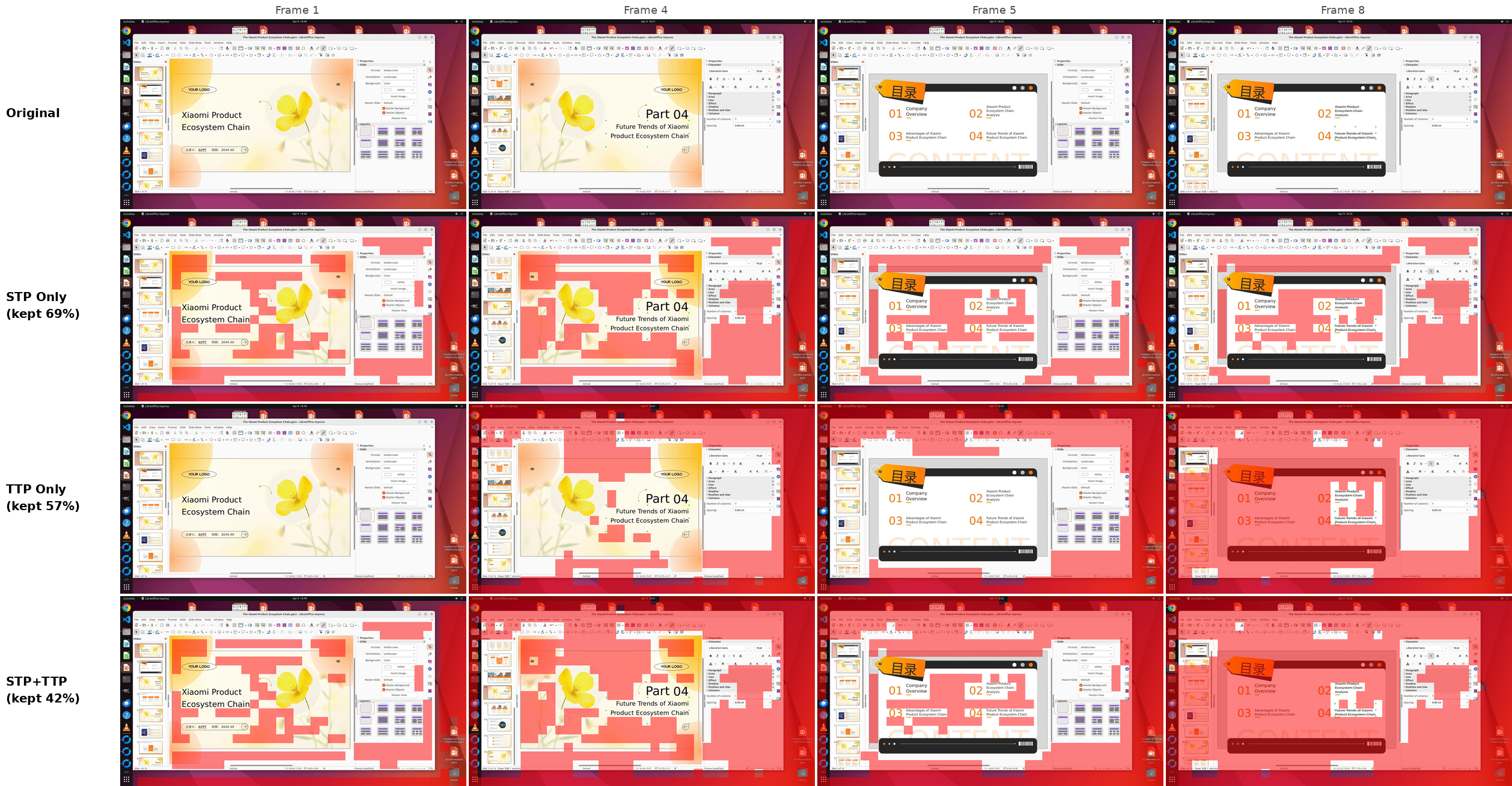}
    \caption{Comparison of STP and TTP}
    \label{fig:stp-ttp-overall}
\end{figure}
\section{Visualization of STP and TTP}
\label{apdx:stp-ttp-visualize}
\autoref{fig:stp-ttp-overall} visualizes how STP and TTP reduce redundancy in video tokens. Under Qwen3-VL tokenization, each frame is represented by 880 tokens after spatial patch merging, and adjacent frames are further merged temporally. To stay consistent with this tokenization pipeline, we first compute STP masks on the original spatial patch grid, and then merge masks for temporally adjacent frames with a logical OR before applying them to merged tokens. This implementation detail explains the blank area on the right side of the first panel in frame 1: if a region is pruned in either of the two frames that are temporally merged, the corresponding merged token remains masked. TTP is then applied directly in token space. Because each token already carries information from neighboring frames, some content that appears early in the trajectory can remain visible in later visualizations. Nevertheless, TTP still removes a large amount of persistent redundancy, including repeated sidebars, window chrome, and static wallpaper/background regions.

This behavior directly matches the motivation of our paper. In \dataset, trajectories are long and highly redundant, while success/failure often depends on subtle local evidence (e.g., small text edits, focus changes, or transient dialogs). STP is effective at filtering large homogeneous regions, but becomes relatively conservative in UI-dense scenes where many small elements may all appear informative. TTP complements this weakness by removing temporally invariant tokens and preserving regions associated with state transitions. Their combination yields a more compact yet task-relevant video representation for reward modeling, which is consistent with the strong \testset performance of \model and its improved temporal attribution compared with strong proprietary baselines.

\section{Prompts}
Here, we present the prompts for the evaluation strategies employed in \textcolor{red}{Findings}~\ref{subsubsec:findings1}. AER~\cite{AER} performs evaluation using only the final screenshot and a compact list of action history, whereas the Simplified Judge~\cite{simplifiedjudge} utilizes both the initial and final screenshots combined with a detailed action history sequence, which covers the action and a longer reasoning of the GUI agent at each step. Although both methods were originally designed for web agent evaluation, we have adapted their prompts to accommodate a wider range of GUI agent tasks. In contrast, SE-WSM~\cite{sun2025seagentselfevolvingcomputeruse} and zeroGUI~\cite{yang2025zeroguiautomatingonlinegui} leverage the full sequence of screenshots without an action history sequence, with the latter operating in a one-shot manner. The prompt for SE-WSM~\cite{sun2025seagentselfevolvingcomputeruse} and ZeroGUI are derived their original paper.

\begin{figure*}[ht]
\centering
\begin{tcolorbox}[
    colback=white,
    colframe=black,
    title={User Prompt of AER~\cite{AER}},
    fonttitle=\bfseries,
    sharp corners,
    boxrule=0.5pt,
    width=\textwidth
]

User Intent:  \verb|{<instruction>}|

Action History:  \verb|{<action history>}|

The last snapshot of the web page is shown in the image.

\end{tcolorbox}
\caption{User prompt of AER~\cite{AER}}
\label{fig:user_prompt_aer}
\end{figure*}

\clearpage
\begin{figure*}[ht]
\centering
\begin{tcolorbox}[
    colback=white,
    colframe=black,
    title={System Prompt of AER~\cite{AER}},
    fonttitle=\bfseries,
    sharp corners,
    boxrule=0.5pt,
    width=\textwidth
]

You are an expert in evaluating the performance of a GUI agent. The agent is designed to help a human user complete tasks across desktop applications and web pages. Given the user's intent, the agent's action history, the final state of the relevant application or webpage, and the agent's response to the user, your goal is to decide whether the agent's execution is successful or not.
\par\vspace{\baselineskip}
There are several types of tasks (not limited to the list below):

1. Information seeking: The user wants to obtain certain information from a webpage, document, or application (e.g., product info, map routes, file properties, system info). The agent's response must contain the requested information or explicitly state that it is not available. If the agent encounters an exception and returns error content, the task is a failure.

2. Navigation: The user wants to navigate to a specific page, view, folder, or screen within a website or application. Carefully examine the action history and final state to determine success. No need to consider the agent's response.

3. Content modification: The user wants to modify content in a webpage, document, or application (e.g., edit text, change formatting, update settings). Verify the final state reflects the requested changes. No need to consider the agent's response.

4. File operations: The user wants to manage files or folders (create, rename, move, copy, delete, compress/extract). Confirm the filesystem state matches the request.

5. Application setup/installation: The user wants to install, uninstall, or configure software, extensions, or plugins. Confirm the final state indicates the requested installation/configuration is complete.

6. Media actions: The user wants to open, play, pause, record, or manipulate images/audio/video. Confirm the media state reflects the requested action.

7. Communication: The user wants to read/send emails or messages. Confirm the content is correctly composed/sent or retrieved.

8. Terminal/command execution: The user wants to run shell commands. Verify commands were executed correctly and outputs match the intent.
\par\vspace{\baselineskip}
Be careful about the sufficiency of the agent's actions. For example, when asked to list the top-searched items, the agent should order items by search count and return the top entries. If required actions (sorting, filtering, saving) are missing, the task is likely a failure.
\par\vspace{\baselineskip}
*IMPORTANT* Format your response into two lines as shown below:

Thoughts: <your thoughts and reasoning process>

Status: "success" or "failure"

\end{tcolorbox}
\caption{System prompt of AER~\cite{AER}}
\label{fig:system_prompt_aer}
\end{figure*}

\clearpage

\begin{figure*}[ht]
\centering
\begin{tcolorbox}[
    colback=white,
    colframe=black,
    title={User Prompt of Simplified Judge~\cite{simplifiedjudge}},
    fonttitle=\bfseries,
    sharp corners,
    boxrule=0.5pt,
    width=\textwidth
]

The user goal is: \verb|{goal}|

The agent performed the following actions:

-----

Step: \verb|{step_number}|

Action: \verb|{action}|

Reasoning: \verb|{reasoning}|

-----

...

-----

Here is the screenshot of the last step.
\verb|{screenshot}|

Provide your reasoning and answer the four questions from the system prompt, usingthe specified format.

\end{tcolorbox}
\caption{User prompt of Simplified Judge~\cite{simplifiedjudge}}
\label{fig:user_prompt_simplified_judge}
\end{figure*}

\clearpage
\begin{figure*}[ht]
\centering
\begin{tcolorbox}[
    colback=white,
    colframe=black,
    title={System Prompt of Simplified Judge~\cite{simplifiedjudge}},
    fonttitle=\bfseries,
    sharp corners,
    boxrule=0.5pt,
    width=\textwidth
]

You are a Language Model specialized in judging the performance of GUI agents in automated interface control tasks. For a given software environment, you are provided with the task goal, the current interface state, the sequence of actions taken by the agent, and the agent's thought process. Additionally, you will have access to the first and last screenshots. Your task is to answer several questions about the agent's performance in the task.
\par\vspace{\baselineskip}
Question 1: Was the sequence of actions below successful in achieving the goal?

Choices: <success>Successful</success>, <success>Unsuccessful</success>
\par\vspace{\baselineskip}
Question 2: Did the agent perform unnecessary actions that could lead to unintended side effects?

Choices: <side>Yes</side>, <side>No</side>
\par\vspace{\baselineskip}
Question 3: Did the agent perform the task optimally, by only performing necessary actions and avoiding unnecessary ones?

Choices: <optimal>1. Complete Failure</optimal>, <optimal>2. Suboptimal</optimal>,
<optimal>3. Somewhat Optimal</optimal>, <optimal>4. Completely Optimal</optimal>
\par\vspace{\baselineskip}
Question 4: Did the agent loop through a sequence of actions that did not make progress towards the goal?

Choices: <loop>Yes</loop>, <loop>No</loop>
\par\vspace{\baselineskip}
You should provide your reasoning process for each question, and you can refer to the screenshots to support your answers. Your answer must follow the following format:

<reasoning>your reasoning here</reasoning>

<success>answer</success>

<side>answer</side>

<optimal>answer</optimal>

<loop>answer</loop>

\end{tcolorbox}
\caption{System prompt of Simplified Judge~\cite{simplifiedjudge}}
\label{fig:system_prompt_simplified_judge}
\end{figure*}

\clearpage
\begin{figure*}[ht]
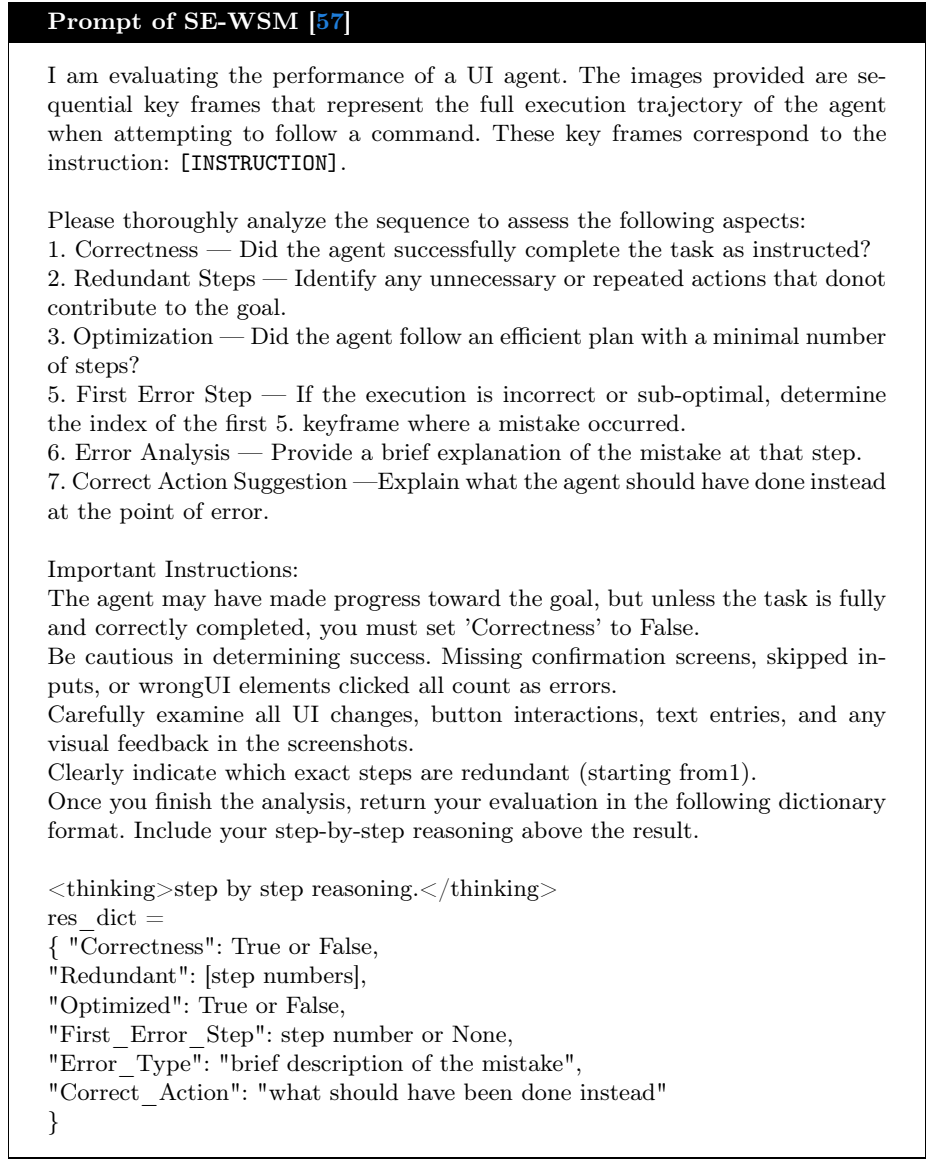

\centering
\begin{tcolorbox}[
    colback=white,
    colframe=black,
    title={Prompt of SE-WSM~\cite{sun2025seagentselfevolvingcomputeruse}},
    fonttitle=\bfseries,
    fontupper=\small,
    sharp corners,
    boxrule=0.5pt,
    width=\textwidth
]

I am evaluating the performance of a UI agent. The images provided are sequential key frames that represent the full execution trajectory of the agent when attempting to follow a command. These key frames correspond to the instruction: \verb|[INSTRUCTION]|. 
\par\vspace{\baselineskip}
Please thoroughly analyze the sequence to assess the following aspects:\\
1. Correctness — Did the agent successfully complete the task as instructed?\\
2. Redundant Steps — Identify any unnecessary or repeated actions that donot contribute to the goal. \\
3. Optimization — Did the agent follow an efficient plan with a minimal number of steps?\\
5. First Error Step — If the execution is incorrect or sub-optimal, determine the index of the first 5. keyframe where a mistake occurred. \\
6. Error Analysis — Provide a brief explanation of the mistake at that step. \\
7. Correct Action Suggestion —Explain what the agent should have done instead at the point of error. 
\par\vspace{\baselineskip}
Important Instructions:\\
The agent may have made progress toward the goal, but unless the task is fully and correctly completed, you must set 'Correctness' to False. \\
Be cautious in determining success. Missing confirmation screens, skipped inputs, or wrongUI elements clicked all count as errors. \\
Carefully examine all UI changes, button interactions, text entries, and any visual feedback in the screenshots. \\
Clearly indicate which exact steps are redundant (starting from1). \\
Once you finish the analysis, return your evaluation in the following dictionary format. Include your step-by-step reasoning above the result. 
\par\vspace{\baselineskip}
<thinking>step by step reasoning.</thinking>

res\_dict = \\
\{ "Correctness": True or False, \\
  "Redundant": [step numbers], \\
  "Optimized": True or False, \\
  "First\_Error\_Step": step number or None, \\
  "Error\_Type": "brief description of the mistake", \\
  "Correct\_Action": "what should have been done instead" \\
\}

\end{tcolorbox}
\caption{Prompt of SE-WSM~\cite{sun2025seagentselfevolvingcomputeruse}}
\label{fig:prompt_SE-ESM}
\end{figure*}

\clearpage

\begin{tcolorbox}[
    colback=white,
    colframe=black,
    title={Prompt of ZeroGUI~\cite{yang2025zeroguiautomatingonlinegui}},
    fonttitle=\bfseries,
    fontupper=\small,
    sharp corners,
    boxrule=0.5pt,
    width=\textwidth,
    breakable,
    enhanced,
]

You will be given a task instruction and a series of screenshots of the task execution.\\
Please analyze the screenshots and provide a detailed analysis of the task completion by following the steps below:\\
1. First, analyze and understand the task instruction. Describe what should the screenshots look like if the task is completed successfully.\\
2. Describe what you observe in each screenshot, analysis what actions were taken and what changes were made to the UI to achieve the task (or mistakes made).\\
3. When you analyze the screenshots, please pay attention to the very detailed elements and changes in the UI. Every small detail may affect the final result.\\
4. After all screenshots are analyzed, provide a overall reasoning about how the task was completed or failed at **the final state**. Make sure you have considered all demands of the task instruction.\\
5. Determine if the task was completed at **the final state** (the last screenshot) successfully (score 1 for success, 0 for failure). If the task is
completed during the process but not at the final state, it should be considered as failure (0 score).
\par\vspace{\baselineskip}
Provide your response strictly in the following format:\\
TASK REQUIREMENT:\\
\lbrack{}Your understanding of the task instruction\rbrack{}
\par\vspace{\baselineskip}
SCREENSHOT ANALYSIS:\\
Screenshot 1:\\
\lbrack{}Analysis of first screenshot\rbrack{}\\
Screenshot 2:\\
\lbrack{}Analysis of second screenshot\rbrack{}\\
...
\par\vspace{\baselineskip}
REASONING:\\
\lbrack{}Your reasoning\rbrack{}

FINAL ANSWER:\\
\lbrack{}Your final answer\rbrack{}

SCORE: \lbrack{}0/1\rbrack{}
\par\vspace{\baselineskip}
Here is an example:\\
(Task Instruction: Please help me backup my emails in "Bills" folder in Thunderbird and store the .eml files with only subject names to my Google Drive folder called "emails".)
\par\vspace{\baselineskip}
TASK REQUIREMENT:\\
- Backup the emails in "Bills" folder in Thunderbird.\\
- Store the backup .eml files with only subject names, and the emails should be saved in the Google Drive folder called "emails". \\
- Once succeed, the emails should be visible in the Google Drive folder "emails". Or at least there should be a saving action performed.
\par\vspace{\baselineskip}
SCREENSHOT ANALYSIS:
\par\vspace{\baselineskip}
Screenshot 1:\\
- Thunderbird email client is open.\\
- The "Bills" folder is visible under "Local Folders."\\
- There is no observable action performed yet in this screenshot.
\par\vspace{\baselineskip}
Screenshot 2:\\
- The "Bills" folder has been selected, and the folder content is displayed.\\
- Two emails are visible: "Amazon Web Services Invoice Available" and "Your receipt from X (formerly Twitter)."\\
- No further actions are taken on the emails.
\par\vspace{\baselineskip}
Screenshot 3:\\
- Both emails in the "Bills" folder are selected.\\
- Content previews of both emails are displayed on the right-hand side.\\
- No observable attempt to export or save the emails is visible.
\par\vspace{\baselineskip}
Screenshot 4:\\
- The right-click context menu is accessed for the selected emails.\\
- The "Save As..." option is hovered over, indicating intent to save the selected emails.
\par\vspace{\baselineskip}
Screenshot 5:\\
- The file navigation window opens, allowing the user to choose a save destination.\\
- No specific Google Drive folder (e.g., "emails") is accessed or visible in this screenshot.
\par\vspace{\baselineskip}
Screenshot 6:\\
- The "Desktop" option in the file picker is hovered over.\\
- Still no indication of Google Drive folder ("emails") selection.
\par\vspace{\baselineskip}
Screenshot 7:\\
- The "Show other locations" option is hovered over in the file picker.\\
- No confirmation that the user is navigating to Google Drive or saving the files with subject names only.
\par\vspace{\baselineskip}
Screenshot 8:\\
- The "Software Updates Available" notification appears. The file picker is still open without any observable confirmation of file saving or destination
selection.\\
- It remains unclear where or if the emails have been saved.
\par\vspace{\baselineskip}
REASONING:\\
Based on the screenshots provided:\\
1. While there was some intent to save the emails (as shown by the selection and access of the "Save As..." function), there is no confirmation that the .eml
files were saved with subject names only and placed in the required Google Drive folder ("emails").\\
2. The screenshots lack evidence of the completion of the task as per the instructions.
\par\vspace{\baselineskip}
FINAL ANSWER:\\
The task was not completed successfully due to the lack of observable saving action.
\par\vspace{\baselineskip}
SCORE: 0
\par\vspace{\baselineskip}
Now, please **strictly follow the format** and analyze the following screenshots (The last line should only be SCORE: [0/1], no other text):\\
Task Instruction: \verb|{instruction}| \\
Screenshots (by order): \verb|{screenshots}|

\end{tcolorbox}
\captionof{figure}{Prompt of ZeroGUI~\cite{yang2025zeroguiautomatingonlinegui}}
\label{fig:prompt_zerogui}

\end{document}